\documentclass[journal]{IEEEtran}

\ifCLASSINFOpdf
\else
\fi
\usepackage[linesnumbered, ruled]{algorithm2e}
\usepackage{setspace}

\usepackage{amsmath}
\usepackage{graphicx}
\usepackage{times}
\usepackage{epsfig}
\usepackage{amssymb}
\usepackage{mathrsfs}
\usepackage[table,xcdraw]{xcolor}
\usepackage{color}
\usepackage[hidelinks]{hyperref}
\usepackage{url}
\usepackage{siunitx}
\usepackage{booktabs}
\usepackage{tabularx}
\usepackage{float}
\usepackage{multirow}
\usepackage{threeparttable}
\usepackage{diagbox}
\newcommand{\revise}{}
\def\degree{${}^{\circ}$}
\usepackage{tikz}
\newtheorem{definition}{\textbf{Definition}}
\newcommand{\rmnum}[1]{\romannumeral #1}
\def \we {HBS }
\def\degree{${}^{\circ}$}
\newcommand{\tabincell}[2]{
\begin{tabular}{@{}#1@{}}#2\end{tabular}
}

\ifCLASSINFOpdf

\else

\fi

\begin{document}
	\title{A Systematic Point Cloud Edge Detection Framework for Automatic Aircraft Skin Milling}

	\author{Zijie Wu,~Yaonan Wang,~He Xie,~Mingtao Feng,~Haotian Wu,~Chao Ding, and Ajmal Mian}
	
	\maketitle
	
	\begin{abstract}
    The edge detection technique is an essential step for aircraft skin milling in aviation manufacturing. 
	Most of the current detection methods focus on traditionally defined edge extraction tasks but disregard the crucial systematic requirement of edge milling. 
	In this paper, we proposed a novel edge detection framework for automatic edge milling of aircraft skins. 
	First, an edge probability detector is proposed by the spatial tangent continuity (STC) to provide the essential reference.
	Second, we propose a hierarchical branch searching (HBS) method to hierarchically strip the desired milling edges from the raw point cloud, which consists of three graded progressive steps: branch backbone generation, branch extension, and branch pruning.
	We demonstrate the performance of the proposed method on both synthetic models and aircraft skin workpieces. The proposed method outperforms the other baselines and shows accurate edges for the edge milling task.
	
	\end{abstract}
	
	

	%
	\IEEEpeerreviewmaketitle

	\section{Introduction}
	\IEEEPARstart{A}{ircraft} skin~\cite{as1} has a significant role in aviation manufacturing and is usually assembled from multiple parts to cover much of the aircraft. The skin parts are riveted to the beams and trusses of the aircraft skeleton. Then, the clearances of adjacent skins are controlled by the edge machining procedure. In the assembly, the edge of the skin part is the critical reference for machining the baseline of the edge milling and the related drilling riveting. More importantly, the milling accuracy of the aircraft skin involves the aircraft aerodynamic performance and stealth performance and is essential for the internal structure safety of the aircraft system.
	The traditional method of aircraft skin machining, including manually comparing and marking the assembly location and removing the edge allowance of the skin part, depends on manual intervention. However, limited by the performance of manual machining, problems including low efficiency, poor consistency, and poor accuracy make it challenging to achieve efficient and accurate aircraft skin assembly.
	
	Motivated by efficient and accurate measurement, the vision-based, 3D measurement system~\cite{tii3}~\cite{feng2020point} is usually more consistent, accurate, and efficient than the traditional manual method and hence applicable to provide a better solution to detect and locate the edge of the skin part. 
	Different from 2D images, point cloud data directly reflect the spatial location of the measured workpiece and are easily utilized for the machining path planning of intelligent manufacturing systems. 
	Nonetheless, a vision-based milling system demands an efficient and accurate edge detection method to separate the desired milling edge from the raw data. Moreover, the thickness of aircraft skins creates difficulties when multiple edge types exist in the current 3D milling edge detection task. 
	Sequentially, it is crucial for robotic edge milling to construct an efficient and accurate milling edge detection framework.

		\begin{figure*}[t!] 
		\center{\begin{tikzpicture}[inner sep=0pt,outer sep=0pt]
		\node[anchor=south west] (A) at (0in,0in)
		{\includegraphics[width=.9\textwidth,clip=false]{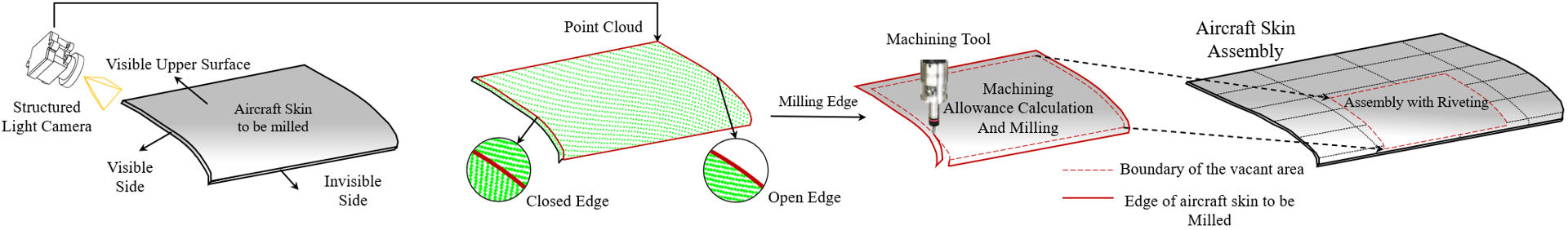}};
		%
		%
		%
		%

		\end{tikzpicture}}
		\vspace{-3mm}
		\caption{\label{fig:skin_assembly} Illustration of the significance of milling edge detection. The point cloud is scanned by the camera and then employed for milling edge detection. The edge information is then applied in robotic edge milling by a machining tool to remove the allowance for final aircraft skin assembly.}
			\vspace{-3mm}
	\end{figure*}

	Fig.~\ref{fig:skin_assembly} shows a common illustration of the assembled aircraft skin. The different viewpoint results in the visible side of thickness and invisible thickness. We define two types of edges in the second part of Fig.~\ref{fig:skin_assembly}, which constitute the desired upper milling edges: 1) A closed edge is usually the intersecting line between two adjacent surfaces. These edges usually have severe curvature variance. 2) An open edge is usually the intersecting line between the measured surface and the blank space.
	The aircraft skin is scanned from the measured surface to the invisible side, and there is no curvature variance but a distinct different point distribution. Because of the complex industrial measurement environment, a complete point cloud is difficult or impossible to obtain in practice, which leads to vague detection caused by the simultaneous appearance of two types of edges. 
	In this paper, we proposed a novel milling edge detection framework based on 3D unorganized point clouds to obtain an accurate milling edge.

	Over the past few decades, many algorithms have been proposed to detect crucial edges for variant applications. Some works~\cite{Canny}~\cite{Sobel} focused on differential operators to detect the gradient variance in 2D pixel images. The Laplacian operator~\cite{Laplacian} detects the image edges by the changes in the second derivative. These methods usually work in most cases and give a fine performance of detected 2D edges. However, these pixel-based, 2D edge detectors~\cite{tii2} highly depend on the texture of the gray level image, and the accuracy is limited by the image resolution. Therefore, subpixel-based detection techniques~\cite{tii1} are proposed to improve the accuracy of edge detection. Zhang~\textit{et al.}~\cite{sbp1} applied the pseudo-Zernike moment to detect the subpixel edges as a preprocessing step. Xiong~\textit{et al.}~\cite{sbp2} combined cubic spline interpolation with mathematical morphology to enhance the reliability of edge detection. Fitting the edge curves~~\cite{sbp3} was proposed to directly improve the accuracy by moving least squares. The subpixel-based detection methods showed better accuracy than pixel-based methods. However, these methods still limit 2D images and cannot achieve the precise edge detection applied for industrial aircraft skin assembly. 
	
	Hence, numerous 3D-based detection techniques have been proposed to address this problem. 
	\emph{Open Edge Detection:} Open Edge detection is usually applied in border extraction between the object and the background.
	One type of solution~\cite{3DLap}~\cite{range1}~\cite{range2_NGRA} in 3D is often an extension of 2D edge detection techniques. Xi~\textit{et al.}~\cite{3DLap} converted point cloud patches that were divided by the coplanar condition to a 2D range image and applied the Laplacian operator to detect the edges. 
	Bao~\textit{et al.}~\cite{range1} used the 2D Canny operator to detect the edges on the range image and included the 3D points corresponding to the detected 2D pixels as the 3D edges. This approach takes advantage of the 2D Canny, which is well applied to 3D areas and shows more robust edge detection performance than other 3D extensions from 2D areas. Steder~\textit{et al.}~\cite{range2_NGRA} proposed the normal aligned radial feature (NARF) descriptor to extract the boundaries (edges). This descriptor explicitly considers the borders of the objects and the backgrounds, which can obtain an accurate result for object feature description. Ahmed~\textit{et al.}~\cite{d2} employed the nonparametric estimation of the mean-shift algorithm by computing the moving distance to classify the edges.
	\emph{Closed Edge Detection:} Closed Edge Detection is often utilized in model contour description to simplify the object shapes.
	Several works focus on the direct detection of 3D point cloud data. Bazazian~\textit{et al.}~\cite{d0} computed the curvature of the local surface by eigenvalue filtering. Although it gives a coarse edge detection result, eigenvalue filtering efficiently utilizes singular value decomposition. Demarsin~\textit{et al.}~\cite{d1} proposed a graph theory to refine the detected edges from normal estimation. The graph-based method~\cite{tii4} is well applied to closed edge detection. Furthermore, the methods using curvature information such as normal are usually time-consuming with low productivity and cannot simultaneously detect all types of edges such as open edges and closed edges. It is not applicable in aviation manufacturing with the demand for high accuracy.
	
	The following two key conditions are significant for achieving accurate and efficient edge detection in industrial manufacturing of aircraft skin assemblies:
	
	1) \emph{The detected edge point should have high accuracy:} The edge milling for aircraft skin assembly requires an accuracy of $\pm$0.5 mm, a thickness of approximately 2-6 mm~\cite{as}, and a longitudinal deflection of $\leq$0.3 mm. The edge information is an important reference for controlling the clearances of adjacent skins and measuring the thickness, so it is necessary for the detection algorithm to output the precise point set of edges. Nonetheless, the features of different types of edges make accurate edge detection challenging.	
	
	2) \emph{Only the desired milling edges need to be detected} In industrial measurement, the closed edges brought by the thickness of the aircraft skins are mixed with the open edges. However, the desired milling edge consists of both types, and hence, the greatest challenge is to accurately detect the milling edge branches to direct robotic milling.
			

	In this paper, we proposed a novel edge detection framework that consists of an edge probability detector using spatial tangent continuity (STC) for edge point description and the hierarchical branch searching (HBS) method to perform efficient milling edge detection on aircraft skins. 
	This method provides a hierarchical branch searching framework for upper milling edge detection on aircraft skins, which solves the three main problems of edge detection described above in aviation manufacturing. In summary, our contributions are given as follows:
	
		1) \emph{A simple continuous geometric feature is proposed to precisely detect edges:} We propose a spatial tangent continuity detector method around the local identified point neighbors to describe the edge probability for each point. We give the common definition of edge on the measured point cloud. The proposed detector can detect the desired edges used for aircraft skin milling.
		
		2) \emph{The hierarchical branch searching framework is proposed for accurate edge detection} We propose a unified edge detection framework that accurately detects the milling edges and approaches a smooth machining path for robotic measurement. This framework is an essential support for the robotic assembly of aircraft skins.


    \section{\label{II-X}Systematic assumptions}
\begin{figure}[t!] 
	\centering{\begin{tikzpicture}[inner sep=0pt,outer sep=0pt]
		\node[anchor=south west] (A) at (0in,0in)
		{\includegraphics[width=.45\textwidth,clip=false]{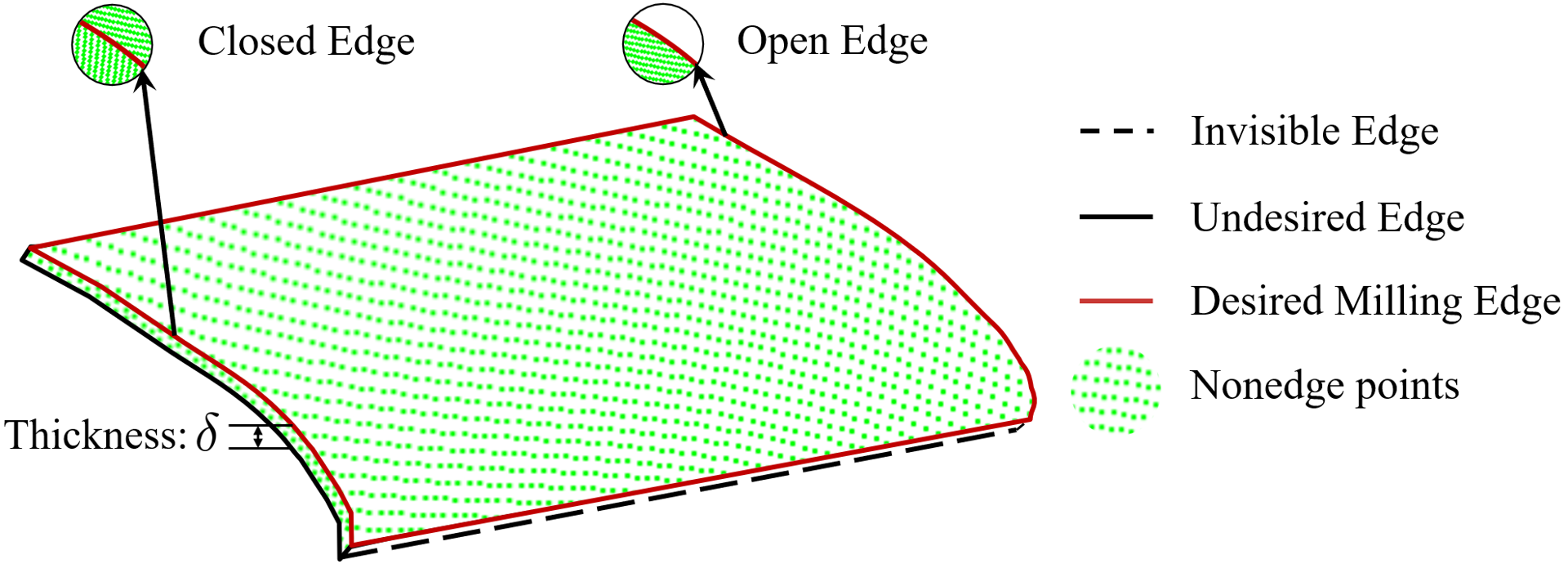}};

		\end{tikzpicture}}

	\vspace{-3mm}
	\caption{\label{fig:pf} Illustration of the systematic problem formulation: the scanned point cloud from the measured aircraft skin consists of multiple types of points.}
	\vspace{-5mm}
\end{figure}

    Aircraft skin edge detection is the problem of detecting the complete milling edges within various perturbations. We consider the point cloud acquisition process from the measured aircraft skin for illustration. As shown in Figure.~\ref{fig:pf}, multiple types of points are acquired with the structure light camera. The milling edge (Solid red line) is defined as the complete and end-to-end boundary for the use of robotic machining requirements.
    The purpose of our edge detection method is to strip the desired milling edge from the scanned raw point cloud. However, various perturbations to the desired milling edge are challenging to achieve accurate detection.

    Several \textbf{assumptions} are considered in terms of the workpiece, point cloud, and systematic process in our detection system:
    \textbf{\emph{Workpiece assumptions:}} 1) The aircraft skin is assumed as the curved workpiece with thickness of $\delta$, which results in multiple types of perturbation to desired milling edge detection. 2) The reflectivity of each measured point on skin is considered to be 100\% scanned. We do not consider the occlusion brought by the low reflectivity of overexposure.
    \textbf{\emph{Point cloud assumptions:}} 3) The scanned point cloud of skins is assumed to be acquired with uniform density. The unusual density points are pre-processed by spatial filtering. 4) Each point cloud contains the location of a crucial physical boundary. Otherwise, the detected edge is not reliable. For instance, there is no edge to detect on the measured point cloud of the metal ball.
    \textbf{\emph{System assumption:}} 5) The process of the milling edge is assumed to be conducted after the scanning and before the skin machining. Edge detection and point cloud acquisition processes are performed separately.

    Workpiece assumptions show that we focus on the problem of perturbations that are resulted from thickness. 
    Point cloud assumptions define the basic parameters of data with uniform density to ensure the efficiency of parallel computing in each sub-area. System assumption is given to illustrate the application in industrial measurement.
    
    As shown in Figure~\ref{fig:pf}, the numerous nonedge points make up the vast majority of the total point cloud. It is crucial to split the edge points from the numerous points of the skin point cloud. Furthermore, the desired milling edge, which is defined as an end-to-end edge on the same surface, is still mixed with a large number of undesired edge points due to the thickness of workpiece assumption. In fact, the undesired and desired edges are difficult to distinguish because of their similar properties.
    Milling edge can be directly applied as the aircraft skin machining path and hence demands high accuracy. The smoothness of the milling path is a necessary requirement for robotic machining. Therefore, detecting a smooth and accurate milling edge, as shown in the solid red lines of Figure~\ref{fig:pf}, is the purpose of our edge detection framework.

	\section{Approach}
\begin{figure}[t!] 
	\centering{\begin{tikzpicture}[inner sep=0pt,outer sep=0pt]
		\node[anchor=south west] (A) at (0in,0in)
		{\includegraphics[width=.42\textwidth,clip=false]{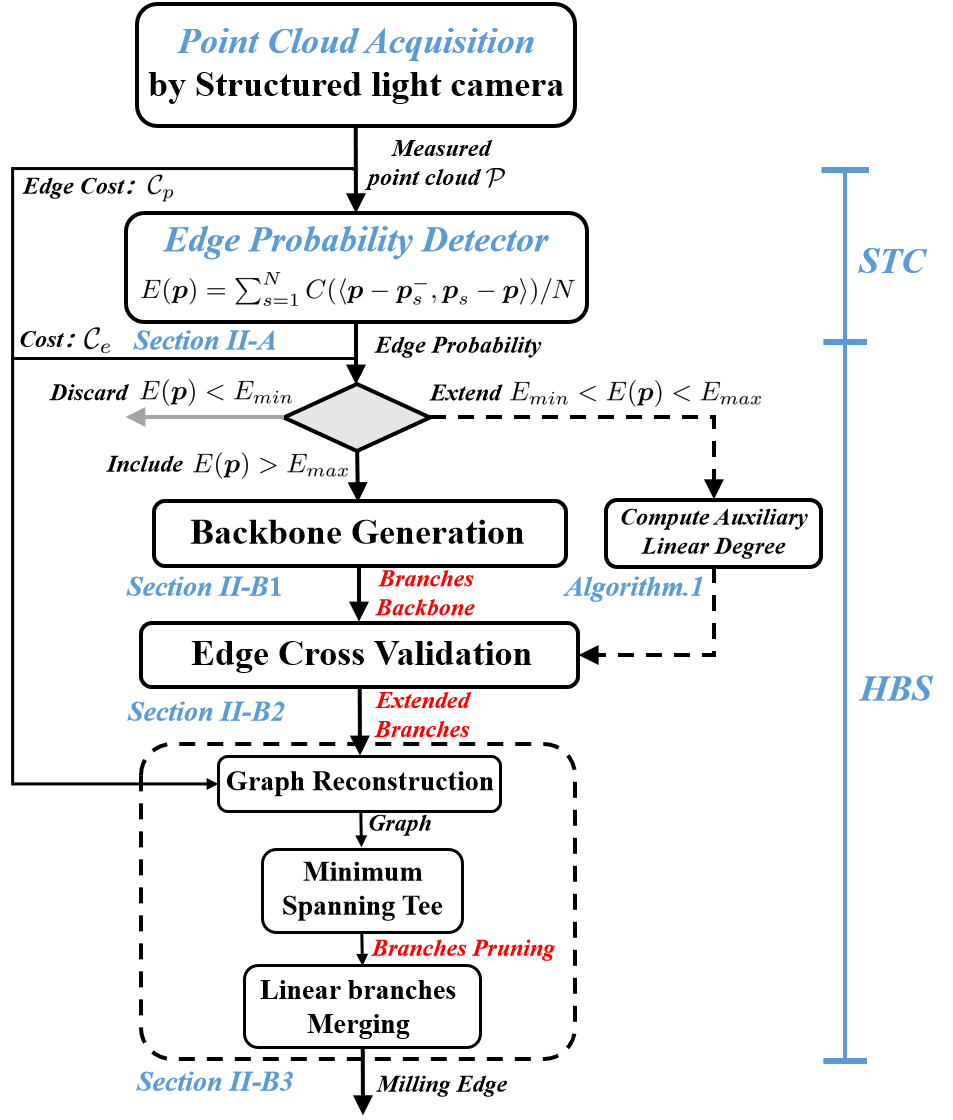}};
		
		\end{tikzpicture}}

	\vspace{-3mm}
	\caption{\label{fig:lct} Framework of the proposed milling edge detection.}
	\vspace{-5mm}
\end{figure}

	In this section, we introduce our edge detection framework, which is shown in Fig.~\ref{fig:lct}. The proposed framework consists of a STC descriptor and a HBS method. STC is computed as the essential reference to instruct edge detection in Section.~\ref{iia}. 
	Based on the proposed STC, we propose the HBS method to hierarchically and accurately detect the desired milling edge in Section.~\ref{iib}.
	First, the edge detector for the initial backbone of edge branch construction is proposed to efficiently establish the general edge structure in Section.~\ref{ii1}.
	Second, based on the features of aircraft skins, we also present a linear evaluation algorithm for more feasible edge point selection upon initial edge selection. This module is designed to smoothly extend the branches from the initial branch backbone. We describe the cross-validation method with the two edge criteria in Section.~\ref{ii2}.
	This process is meant to include the edge points based on the initial structure as much as possible but also introduces the incorrect edge branches that are hard to distinguish. Hence, we proposed a graph optimization for edge refinement in Section~\ref{ii3} to prune these incorrect point branches and to obtain an accurate and smooth edge for robotic aircraft skin milling. The main advantages of the hierarchical branch searching method can be concluded as:\begin{enumerate}
        \item HBS employs a local edge-aware STC descriptor and designs a hierarchical framework to precisely select the edge point in 3D space.
        \item Hierarchical framework is tolerant to the multiple types of perturbations. For instance, the uncompleted bottom edge lines can be recognized and pruned, which is crucial for automatic aircraft skin measurement.
        \item HBS detects an accurate milling edge that can be directly applied for robotic machining.
	\end{enumerate}
	
	\subsection{Spatial Tangent Continuity for Edge Point Selection~\label{iia}}
	
	Different from the existing method that relies on the point distribution, we rethink the definition of the edge and nonedge in 3D. 
	Here, the spatial tangent continuity is proposed to distinguish the nonedge point from edge point. We define the \emph{tangent continuity at a point} condition:
	
	\begin{definition}
		\emph{(discrete tangent continuity for a spatial nonedge point) If point $\boldsymbol{p}$ does not belong to the edge, there exists a constant $\delta>0$ such that all points $\left\|\boldsymbol{p}_{s}-\boldsymbol{p}\right\|<\delta$ of the measured surface constitute a neighbor plane $\boldsymbol{P}$. The plane $\boldsymbol{P}$ is}
		
		\emph{(\rmnum{1})} strictly tangent continuous at $\boldsymbol{p}$, \emph{if $\exists$ $\boldsymbol{p}_{s}^{-}\in \boldsymbol{P}$ and there exists a constant $\varepsilon >0 $ ,} 
		\begin{equation}
			\left<\boldsymbol{p}-\boldsymbol{p}_{s}^{-},\boldsymbol{p}_{s}-\boldsymbol{p}\right> <\varepsilon 
		\end{equation}
		\emph{for all $\boldsymbol{p}_{s} \in \boldsymbol{P}$, $\left<,\right>$ is the angle between two vectors,}
		
		\emph{(\rmnum{2})} strongly tangent continuous at $\boldsymbol{p}$ \emph{if $\exists$ $\boldsymbol{p}_{s}^{-}\in \boldsymbol{P}$,}
		\begin{equation}
		\left<\boldsymbol{p}-\boldsymbol{p}_{s}^{-},\boldsymbol{p}_{s}-\boldsymbol{p}\right> =0
		\end{equation}
		\emph{for all $\boldsymbol{p}_{s} \in \boldsymbol{P}$.}

	\end{definition}

		\begin{figure}[t!] 
		\begin{tikzpicture}[inner sep=0pt,outer sep=0pt]
		\node[anchor=south west] (A) at (0in,0in)
		{\includegraphics[width=.48\textwidth,clip=false]{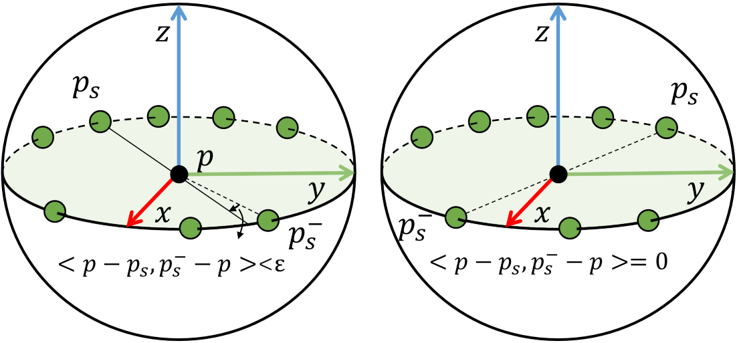}};

		\node[anchor=north,xshift=-63pt,yshift=-2pt] at (A.south)
		{\footnotesize (a) Strictly continuity};
		\node[anchor=north,xshift=63pt,yshift=-2pt] at (A.south)
		{\footnotesize (b) Strongly continuity};
		
		\end{tikzpicture}

		\vspace{-3mm}
		\caption{\label{fig:smt} Illustration for the definition of spatial tangent continuity. }
		\vspace{-5mm}
	\end{figure}

	Definition 1 is illustrated in Fig.~\ref{fig:smt}. If $\boldsymbol{P}$ is strongly tangent continuous at $\boldsymbol{p}$, then $\boldsymbol{P}$ is strictly tangent continuous at $\boldsymbol{p}$. Definition 1(\rmnum{2})~is considered the tangent continuity of all points in the continuous surface, and Definition 1(\rmnum{1}) is the generalized solution upon the point cloud. We apply the pairwise vector to approach the tangent and compute whether a collinear and opposite tangent exists for all tangents that go through the point $\boldsymbol{p}$.
	With the above definition, we obtain the following discrete computation to obtain a probability of the point $\boldsymbol{p}$ that belongs to an edge.

	\textbf{Discrete computation of tangent continuity on the point cloud:}
	The 3D point cloud is considered the observation point set from the measured continuous surface. However, due to the essence of the observation sample, the point cloud is inherently sampled with gaps. We only consider a strict continuity on discrete point cloud data to extract the edge point. 
	
	For a point $\boldsymbol{p}$, we apply discrete computation for continuity estimation. First, we consider the neighbors $\boldsymbol{p}_{s=1}^{n}=\{\boldsymbol{p}_1, \boldsymbol{p}_2,..., \boldsymbol{p}_N\}$ for which the Euclidean distance is less than $\delta$, where $N$ is the number of neighbors. Second, for each neighbor point $\boldsymbol{p}_{s}$, we estimate if it meets the condition of strict continuity. The greater the number of $\boldsymbol{p}_{s}$ that is not strictly continuous, the higher the probability that it belongs to an edge. The probability $E$ is computed as:
	\begin{equation}
		E(\boldsymbol{p})=\sum_{s=1}^{N} C(\left<\boldsymbol{p}-\boldsymbol{p}_{s}^{-},\boldsymbol{p}_{s}-\boldsymbol{p}\right>)/N,
	\end{equation}
	where $C$ is the counting function for the number of $\boldsymbol{p}_{s}$ when there is no $\boldsymbol{p}_{s}^{-}\in \boldsymbol{p}_{s=1}^{n}$ meeting the condition that $\left<\boldsymbol{p}-\boldsymbol{p}_{s}^{-},\boldsymbol{p}_{s}-\boldsymbol{p}\right> <\varepsilon$. 
	\begin{equation}
	C(\left<\boldsymbol{p}-\boldsymbol{p}_{s}^{-},\boldsymbol{p}_{s}-\boldsymbol{p}\right>)=\begin{cases}
	0,&{\rm for} \quad \text{strictly continuity}\\
	1.&{\rm for}\quad else,
	\end{cases}
	\end{equation}
	In practice, the neighbor points are not always distributed within a neighbor plane. Against the numerical sampling radius $R_c$, we apply the adaptive continuity threshold:
	\begin{equation}
		\varepsilon=\frac{2\pi}{N}  \exp^{ \frac{\left\| \boldsymbol{p}-\boldsymbol{p}_{s}  \right\|}    {  R_c}}   ,
	\end{equation}
	This term is applied by considering more tangent bias tolerance as the distance increases.
	Based on the edge probability of each point, point $\boldsymbol{p}$ is selected as an edge candidate by the proposed HBS method. 

	\subsection{\label{iib}Hierarchical Branch Searching for Milling Edge Detection}
	In this section, we propose a hierarchical branch searching method to detect the desired milling edge.
	
	\subsubsection{\label{ii1}\textbf{Branch Backbone Generation}} For aircraft skins, the edge points of complex edges show much divergence from each other. For instance, points in corners, lines, arcs, etc., have different point distributions and result in different edge probabilities.
	
	The edge candidates generated by probability selection contain many suboptimum edge points, which yields a high probability because their locations are near the real edge. To determine the optimal edges, the problem that we faced is to choose the threshold of the probability score for edge point identification. However, due to the various edges, such as convex and concave edges, the confidence probability score of optimal edge points is different for each edge in a point cloud. A single threshold value is difficult to distinguish for efficient edge detection, and a dynamic and flexible filter scheme is needed.
	
	Hence, we divide the measured points with a threshold tuple $\{E_{max},E_{min}\}$.
	We generate a rough branch backbone using high edge probability points by selecting the point $\boldsymbol{p}$ when $E(\boldsymbol{p})$ is larger than the threshold $E_{max}$ and discard the point $\boldsymbol{p}$ when $E(\boldsymbol{p})$ is lower than the threshold $E_{min}$. 
	
	\subsubsection{\label{ii2}\textbf{Branch Extending by Cross Validation}} Based on the edge probability $E(\boldsymbol{p})$, the edges can be coarsely selected by threshold filtering. However, as shown in most edge detection methods, the single threshold filter hardly addresses the complex edge variation, especially on aircraft skins. For point $\boldsymbol{p}$ with $E_{min}\leq E(\boldsymbol{p}) \leq E_{max}$, it is difficult to judge only with the edge probability criterion. We introduce an auxiliary criterion, \emph{Linear Degree}, for better branch extension as follows:
	
	Hence, the linear degree is applied to provide extra information, which is defined as:
	\begin{definition}\label{d2}
		\emph{(Linear Branches Degree)} For an ordered point set $\{\boldsymbol{p}|\boldsymbol{p}_1\rightarrow \boldsymbol{p}_2\rightarrow\boldsymbol{p}_3\rightarrow\boldsymbol{p}_4\rightarrow\cdots\rightarrow\boldsymbol{p}_M\}$, this branch is assumed to have a linear degree if all $\{<\boldsymbol{p}_i-\boldsymbol{p}_{i-1}\rightarrow \boldsymbol{p}_{i+1}-\boldsymbol{p}_{i}>\}\leq \theta$.
		
	\end{definition}
	 
	 The value $\theta=0$ usually determines a straight line. For better tolerance of measurement noise, $\theta$ is usually set as a sufficiently small constant.

	\textbf{Linear Clustering:} A clustering algorithm is proposed to separate the line out; for instance, the lines that belong to edges contain a larger number of points. Here, $\mathcal{P}$ is the candidate set of all selected edge points and $\boldsymbol{p}_{i}\in\mathcal{P}$ is the $i$-th point. 
	We propose the progressive accessible clustering algorithm (PAC) to fulfill the linear degree estimation and implement it based on the backbone of the DBSCAN~\cite{DBSCAN} algorithm.
	
	The PAC is proposed to segment edges from local neighbors. This algorithm is based on the concept of Progressive and Accessible, which is defined as follows: if the neighbor point $\left\| \boldsymbol{p}_A-\boldsymbol{p}_i\right\|<\epsilon_a$, $\boldsymbol{p}_A$ is accessible from $\boldsymbol{p}_i$, each time of accessible propagation, the depth from measured point $\boldsymbol{p}$ to $\boldsymbol{p}_i$ must be increasing. A cluster consists of all segments that are accessible to each other. Furthermore, the clusters whose number is less than $min\_pts$ would be labeled as noise points and deleted from linear estimation. The whole accessible clustering algorithm is detailed in Algorithm.~\ref{algorithm1}. The number of linear clusters provides extra information to detect the edge point when it has a relatively smaller edge probability. 
	
	\begin{algorithm}
		\label{algorithm1}
		\SetAlgoLined
		\KwIn{Clustering Candidates $\mathcal{P}$, Measured point $\boldsymbol{p}$, \\$\quad\qquad$Least cluster number $min\_pts$, \\$\quad\qquad$Accessible radius $\epsilon_a$;}
		\KwOut{Linear Degree $L$;}
		Init: Current cluster label $L=0$, depth $D$=0;\\
		$\mathcal{P}$ is set to the origin: $\mathcal{P}_i=\mathcal{P}_i-\boldsymbol{p}$;\\
		$\mathcal{P}$ is projected to the unit ball: $\mathcal{P}=\mathcal{P}/length(\mathcal{P})$;\\
		\ForEach{unlabeled point $\boldsymbol{p}_i \in \mathcal{P}$}{
			The current depth is updated: $D_{last}=D$,$D=length(\boldsymbol{p}_i)$;\\
			$\mathcal{Q}=RangeSearch(\boldsymbol{p}_i,\epsilon_a)$;\\
			\While{$\mathcal{Q}\neq \varnothing$}{
				\If{$(D-D_{last})\cdot(\left\| \boldsymbol{p}_i\right\|-\left\| \boldsymbol{p}_{i-1}\right\|)$}{Label $\boldsymbol{p}_i \in \mathcal{Q}$ as $L$;}
				$\mathcal{Q}=RangeSearch(\mathcal{Q},\epsilon_a)$;\\

			}

			$L=L+1$;
			
		}	
		\For{$j\in L$}{
			\If{$|\text{Cluster}_j|\leq min\_pts$}{
				$\text{Cluster}_j$ is labelled as noise;
				
			}
		}
		
		\caption{Progressive Accessible Clustering Algorithm}
	\end{algorithm}

	\textbf{\emph{Cross Validation by Auxiliary Linear Degree:}}
	To improve the smoothness of edges, we follow the works of~\cite{Canny}, and the linear degree of the edge point is evaluated to ensure the quality of the generated edges. We employ two thresholds $E_{min}$ and $E_{max}$ ($E_{min}<E_{max}$) in the cross-validation process, where a value lower than $E_{min}$ represents a low confidence of the nonedge point to discard these points and a value higher than $E_{max}$ to include the high confidence points as edges. The backbone of the initial edges is extracted by selecting the high confidence points with the edge probability detector.
    
    As defined in \emph{Definition}.~\ref{d2}, all points of the branch backbone are computed to obtain their related linear branches. These edge points of aircraft skins should have their branch points, which also belong to the edges. Hence, we evaluate both criteria to approach the accurate edges, namely, the cross validation:
	\begin{enumerate}
		\item The linear branches of all points in the initial edges that $E(\boldsymbol{p})>E_{max}$ are computed, and all points that $E(\boldsymbol{p})<E_{min}$ are discarded.
		\item The points in the initial edges that have no linear branches are discarded and labelled as noise.
		\item These points in the linear branches that $E(\boldsymbol{p})>E_{min}$ are included as edges.
	\end{enumerate} 
	Based on this cross-validation process, the edges consisting of points and branches are detected.

	\begin{figure}[t!] 
	\begin{tikzpicture}[inner sep=0pt,outer sep=0pt]
	\node[anchor=south west] (A) at (0in,0in)
	{\includegraphics[width=.48\textwidth,clip=false]{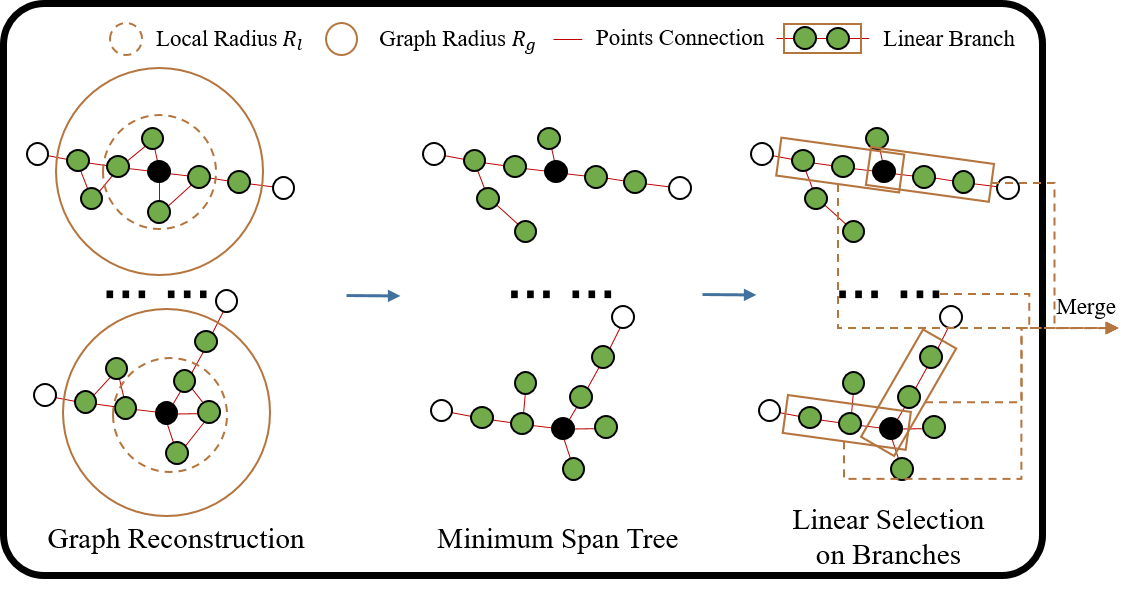}};
	
	\end{tikzpicture}

	\vspace{-3mm}
	\caption{\label{fig:graph} Branch Refinement by Graph Optimization. Every row proceeds in parallel, and all linear branches are merged at the end to constitute the final milling edge.}
\end{figure}
	
	\subsubsection{\label{ii3}\textbf{Branch Pruning by Graph Optimization Refinement}}
	Traditional methods for edge detection end with the edge candidates that are selected in Section~\ref{ii1}. However, industrial manufacturing, such as measuring and machining aircraft skin, demands an accurate 3D case. Vague information usually leads to measurement bias and machining jitter. We propose a branch extension scheme to improve the precision. 
	In this section, we propose a graph optimization theory to guarantee smoothness for better milling edge detection to support the robotic system.

	\textbf{\emph{Graph reconstruction:}} Based on the branches extending after backbone establishment, the milling edge is roughly detected. However, for milling operations, the edge is desired with smoothness and accuracy. Hence, the graph reconstruction method is applied for error branch pruning.
	As shown in Fig~\ref{fig:graph}, we consider the coarse edges after clustering as the backbone of the graph structure. 
	
	$(\rmnum{1})$ All points in a cluster are locally connected to the neighborhood graph $G$ to prevent local perturbations. In practice, the perturbations distort the points around the real edges and cause the incorrect connection if a graph is established with few connected neighbors. To solve this problem, previous works usually connect all points on a large scale, which would be computationally intensive to prune the numerous unwanted connections. Furthermore, the different lines of aircraft skin edges are not always far from each other. The scheme of large-scale connections would produce unexpected connections. In our solution, we give a related feasible receptive field to produce the local connections to enable path searching across the local perturbations. These connections with the local receptive field of radius $R_l$ are constructed on the backbone of the clustered coarse edges to a global graph.

	$(\rmnum{2})$ The local optimal edge of point $\boldsymbol{p}$ is estimated by all its neighborhood connections with graph radius $R_g$ ($R_g>R_l$). The local graph of point $\boldsymbol{p}$ is taken out to search for an ordered optimal path by the shortest path method. Formally, given the chains of local adjacent edges, there is only one path of the shortest path searching by the defined path cost from $\boldsymbol{p}$ to each neighborhood. The path cost is defined in two parts. We obtain the probability cost:
	\begin{equation}
		\mathcal{C}_{p}=1-\frac{E(\boldsymbol{p}_{i})+E(\boldsymbol{p}_{j})}{2},
	\end{equation}
	The cost forces on the connections are obtained from the points with high edge probability. In addition, a Euclidean distance cost is applied to connect the neighbors:
	\begin{equation}
	\mathcal{C}_{e}=1-\frac{d_{\#}}{\left\| \boldsymbol{p}_{i}-\boldsymbol{p}_{j} \right \|},
	\end{equation}
	where the Euclidean cost is to force edges from the nearer neighbors. $d_{\#}$ is the average distance of the measured point cloud.
	
	The overall cost is $\mathcal{C}=\mathcal{C}_p+\mathcal{C}_e$.
    Fig.~\ref{fig:graph} shows the diagram of the graph optimization. We construct our local optimal edge on the connected cost function by running the minimum span tree algorithm~\cite{MST} and by searching the shortest paths~\cite{DJ}.
	Each neighborhood point generates an ordered path from $\boldsymbol{p}$ that is represented by the chains of adjacent edges. All paths from $\boldsymbol{p}$ are evaluated for their linear degree, and hence, the top two nonoverlapping edges are selected based on the linear degree to constitute the upper milling edges. The purpose of the linear degree evaluation on the edge graph instead of directly on the edge points is to better preserve the corner points and improve the accuracy.

    \subsection{\label{iiic} Time Complexity Analysis}
    Assume the point cloud of the measured point cloud contains the $m$ number of total points and the $n$ number of edge points. 
    The proposed approach consists of branch backbone generation, branch extending, and branch pruning.
    In branch backbone generation, the computation of STC for every point consists of a k nearest neighbor (knn) searching for all points and a top-1 nearest neighbor searching in the knn set.
    The construction of the searching tree takes $O(m\log{m})$ time. The top-1 nearest neighbor searching of STC takes $O(\log \mathcal{R})$ time, in which $\mathcal{R}$ is the number of neighbors.

    In branch extending, the time cost of the branches extending process is mainly from the linear clustering scheme using the PAC algorithm and takes $O(n^2)$ time. Using the k-dimension tree building, the time complexity can be reduced to $O(n \log n)$

    Last, for branch pruning, the total time cost consists of minimum span tree construction around the desired edge points and shortest path searching. Minimum span tree construction takes $(O(n)\cdot O({R}^2_g\log R^2_t)$ time, where the ${R}_g,{R}_t$ are the knn searching radius take $2\cdot O(n \log n)$ time. Shortest path searching take nearly $O(n)\cdot O({R}_g^2 \cdot R_t^2)$ time.

    Furthermore, the parallel technique can be used to speed up HBS, since the algorithm is inherently task-parallel in all three processes due to the uniform density of point cloud assumption. Especially when dealing with low canning density, the time complexity of branch pruning can be reduced to $O(\log n)\cdot (O({R}^2_g\log R^2_t) + O({R}^2_g \cdot R^2_t))$.  We abbreviate $O(\log \mathcal{R})$, $(O({R}^2_g\log R^2_t) + O({R}^2_g \cdot R^2_t))$ as $K_1$ and $K_2$ respectively because they can be regarded as a constant when dealing with the uniform density of point clouds.
    
    As a result, the whole edge detection consumes time $T$ to be accomplished, in which $T$ is computed as follows:
    \begin{equation}
    \begin{aligned}
        T&=K_1 O(m\log{m}) + 3\cdot O(n \log n) +K_2 O(\log n)
        \end{aligned}
    \end{equation}

    As the usual condition, the proportion of $n/m$ decreases as the point density increase. For instance, $m\gg n \gg \mathcal{R} \gg R_g >R_t$ when dealing with a large number of points, the time cost is dominated by the STC construction. $T\approx K_1 O(m\log{m})$.

%
%
	
			\begin{figure}[t!] 
		\begin{tikzpicture}[inner sep=0pt,outer sep=0pt]
		\node[anchor=south west] (A) at (0in,0in)
		{\includegraphics[width=.48\textwidth,clip=false]{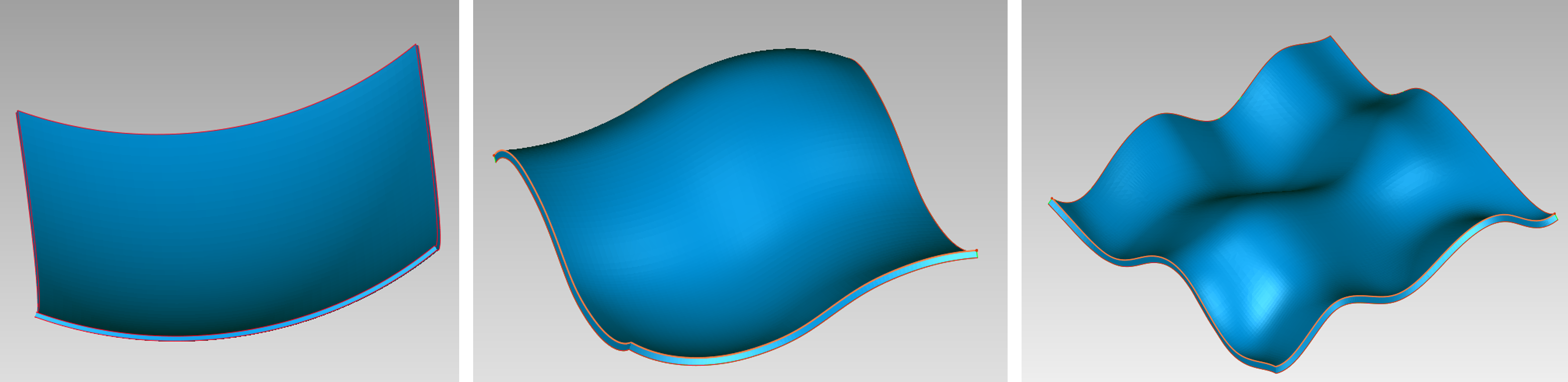}};
			\node[anchor=north,xshift=-85pt,yshift=-2pt] at (A.south)
			{\footnotesize (a) Model 1};
			\node[anchor=north,xshift=-4pt,yshift=-2pt] at (A.south)
			{\footnotesize (b) Model 2};
			\node[anchor=north,xshift=85pt,yshift=-2pt] at (A.south)
			{\footnotesize (c) Model 3};
		
		\end{tikzpicture}

		\vspace{-3mm}
		\caption{\label{fig:model} Skin models}
	\end{figure}
%
%
%
%
%
%

\begin{figure}[htbp] 
\center{	\begin{tikzpicture}[inner sep=0pt,outer sep=0pt]
	\node[anchor=south west] (A) at (0in,0in)
	{\includegraphics[width=.38\textwidth,clip=false]{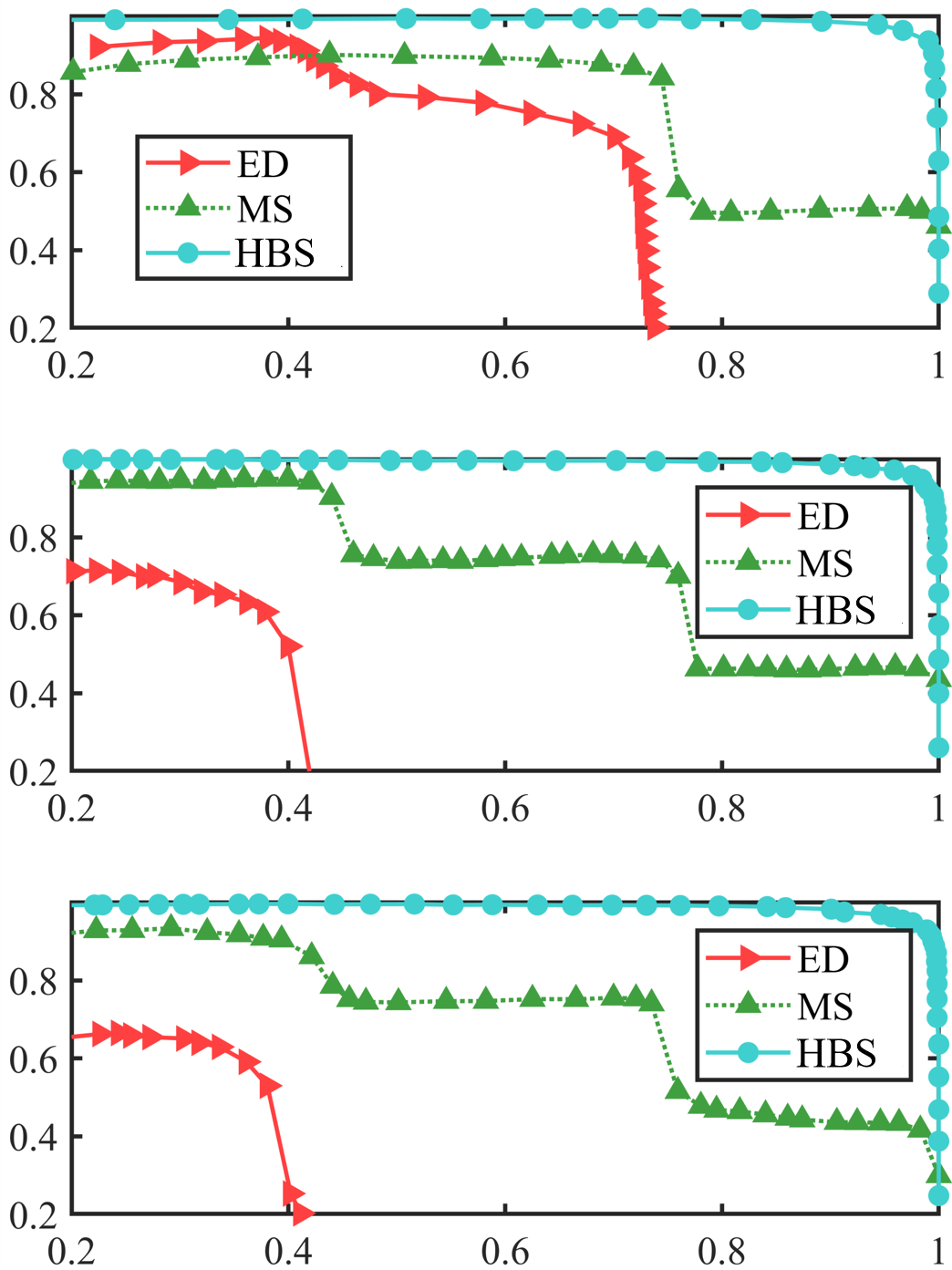}};
	
	\node[anchor=north,xshift=0pt,yshift=0pt] at (A.south)
	{\footnotesize (a) Recall ($\times100\%$)/Model 3};
	\node[anchor=north,xshift=0pt,yshift=90pt] at (A.south)
	{\footnotesize (b) Recall ($\times100\%$)/Model 2};
	\node[anchor=north,xshift=0pt,yshift=180pt] at (A.south)
	{\footnotesize (c) Recall ($\times100\%$)/Model 1};

	\node[anchor=north,xshift=-5pt,yshift=90pt,rotate=90] at (A.west)
	{\footnotesize Precision ($\times100\%$)};
	\node[anchor=north,xshift=-5pt,yshift=0pt,rotate=90] at (A.west)
	{\footnotesize Precision ($\times100\%$)};
	\node[anchor=north,xshift=-5pt,yshift=-90pt,rotate=90] at (A.west)
	{\footnotesize Precision ($\times100\%$)};

	\end{tikzpicture}}

	\vspace{-3mm}
	\caption{\label{fig:P2C} Precision/Recall evaluation of different edge detectors.}
\end{figure}

%
%
%
%
%
%
%
%
	\section{Experimental Results}
	In this section, three parts are designed and dedicated to testing the performance of an aircraft skin model with a complex surface. The first part tests the performance with four metrics to show the effectiveness of our edge detection method in Section A. The second part is designed to show the perturbation resistance ability with respect to measurement noise and outliers in Section B. These previous two parts are conducted based on the synthetic dataset for better evaluation. Hence, we evaluate the edge detection experiment above the standard workpieces on our robotic measurement system in the third part. The third part further verifies the advantages of our proposed method in a practical industrial environment. In addition, an aircraft skin with a complex free-form surface is also included to visualize the extracted edge. All aircraft skins are designed with a thickness of 5$mm$.

	\textbf{Baselines:} We compare our method \we with two typical edge detection methods: ED represents detection via eigenvalue variations, and MS\cite{d2} is a mean-shift edge detector that relies on the neighbor point distribution.
	
	\textbf{Evaluation Metrics:} Here, we employ four metrics~\cite{juba2019precision} to evaluate the quality of edge extraction methods: accuracy, precision, recall, and error. Note that, the error is defined as the distance to the nearest reference points.

%
%

	\subsection{\label{iiia}Evaluation on the performance}
	In this section, we model the aircraft skin measurement with limited views to obtain point cloud data that simultaneously contain the open edges and corner edges. 
	
	\textbf{\emph{1) Robustness and effectiveness evaluation of the proposed algorithm:}} Here, we focus on the performance comparisons to show the robustness and effectiveness of our proposed method using three complex surface models shown in Fig.~\ref{fig:model}. These models are designed with complex free-form surfaces. In addition, both the bottom and two sides, which indicate the thickness of the model, are removed from the surface model to simultaneously create the structure with closed edges and open edges.
	These models are measured with fewer views, which improves the efficiency and accuracy to extract the edge information.
	
	\textbf{Edge Comparison of Performance Evaluation:} In this experiment, we evaluate the quality of the extracted edges upon three complex surface models. Given the CAD models, the model shapes consist of two parts: the nonedge points (90k points), which are directly sampled on the surface and the edge points (3k points) sampled from the edge curves. Because few surface sampling points are close enough to the ground-truth edges, we label these points $\left\| \boldsymbol{P}-\boldsymbol{P}_{gt}\right\|<\frac{1}{10}*d_{res}$ as edge points, even if they do not belong to the edge samples. Here, $d_{res}$ is the average point distance of the current point cloud.
	We conduct experiments to show the performance of the spatial tangent continuity detector with two 3D edge detectors, which are described in the Baselines. Note that for a fair comparison, we only evaluate the original detectors here by disabling the cross-validation and refinement of HBS. Quantitative analysis of the performance results is shown in Fig.~\ref{fig:P2C}, using the precision to recall estimation on the three models. As shown in Fig.~\ref{fig:P2C}, Figures (a)-(c) reflect the performance of Models 1-3, respectively. It is obvious that our detector of HBS has consistently higher precision and recall, which outperforms the other algorithms. Furthermore, ED represents the curve variation detection using eigenvalues and achieves high precision when the recall is low, but it quickly drops when out of range. This situation worsens, especially as the surfaces become more complex, such as when the precision directly falls when the recall is larger than 0.4 on Models 2 and 3 with complex free-form surfaces. 
	The MS has a stepped performance on all figures, which shows that the points are detected with several common features and easily classified as different clusters. The complex shapes on Models 2 and 3 cause more divergence of MS features. Furthermore, as the recall increases, the MS cannot achieve precise edge detection on these complex models.

	In addition, we also show a more quantitative comparison that is detailed in Table~\ref{tab:aL1}. Here, we select one node of the above experiments that follows the max F1-score~\cite{f1s} from each method of model comparison. 
	We employ the F1-score to evaluate the suitable performance of the different detectors and conduct a comparison between the proposed HBS and baselines.
	
	
		\begin{figure*}[htbp] 
	\begin{tikzpicture}[inner sep=0pt,outer sep=0pt]
	\node[anchor=south west] (A) at (0in,0in)
	{\includegraphics[width=.98\textwidth,clip=false]{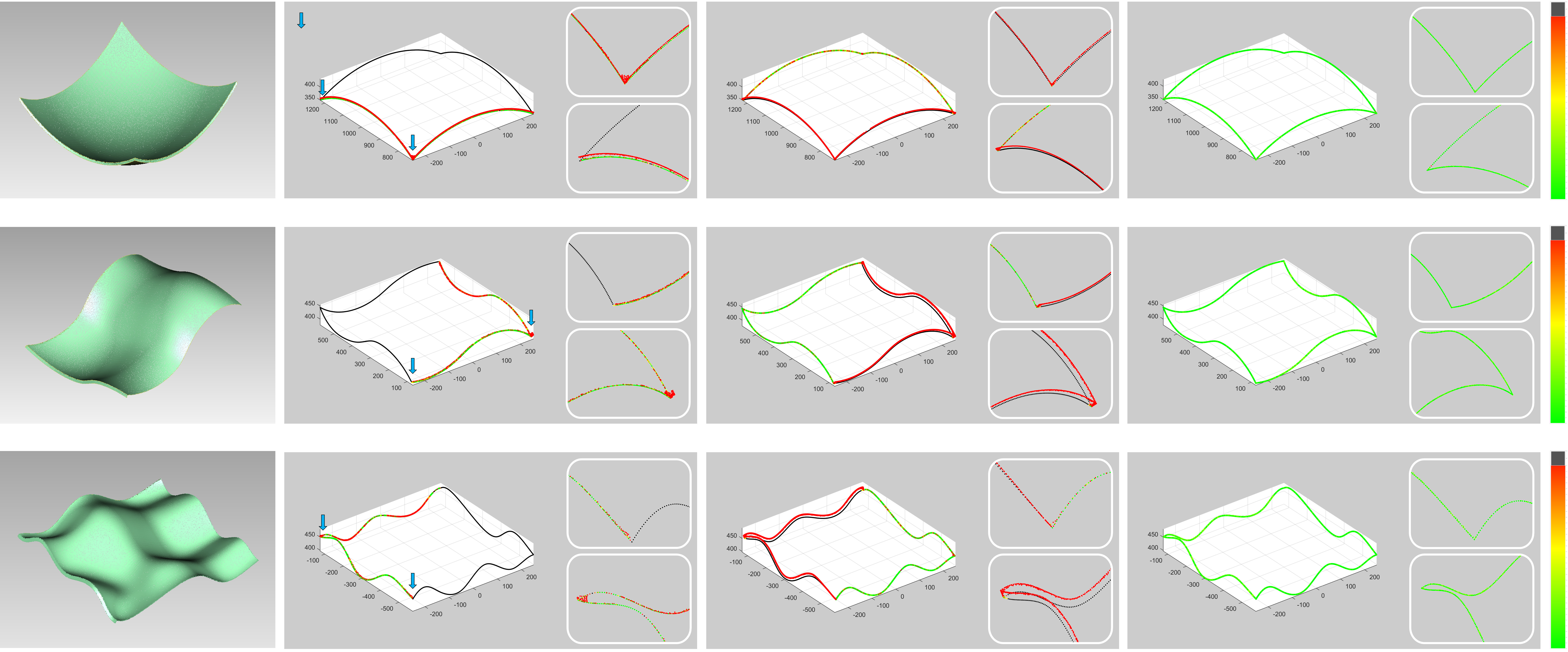}};
	
	\node[anchor=north,xshift=-205-3pt,yshift=-3pt] at (A.south)
	{\footnotesize (a) Sampled point cloud};
	\node[anchor=north,xshift=-106-10pt,yshift=-3pt] at (A.south)
	{\footnotesize (b) ED};
	\node[anchor=north,xshift=-66+60+30pt,yshift=-3pt] at (A.south)
	{\footnotesize (c) MS};

	\node[anchor=north,xshift=185+16-40pt,yshift=-3pt] at (A.south)
	{\footnotesize (d) HBS};
	
	\node[anchor=north,xshift=-106-14.5pt,yshift=-4.5pt] at (A.north)
	{\tiny Location of the enlarged area};

	\node[anchor=north,xshift=-7pt,yshift=0pt,rotate=90] at (A.west)
	{\footnotesize Model 2};
	\node[anchor=north,xshift=-7pt,yshift=70pt,rotate=90] at (A.west)
	{\footnotesize Model 1};
	\node[anchor=north,xshift=-7pt,yshift=-75pt,rotate=90] at (A.west)
	{\footnotesize Model 3};
	
		\node[anchor=north,xshift=6pt,yshift=+38.5pt] at (A.east)
	{\scriptsize [mm]};
		\node[anchor=north,xshift=8pt,yshift=+30pt] at (A.east)
	{\scriptsize 1.00};
		\node[anchor=north,xshift=8pt,yshift=18.4pt] at (A.east)
	{\scriptsize 0.00};
		\node[anchor=north,xshift=8pt,yshift=6.4+0.4pt] at (A.east)
	{\scriptsize 0.60};
		\node[anchor=north,xshift=8pt,yshift=-5.4+0.6pt] at (A.east)
	{\scriptsize 0.40};
		\node[anchor=north,xshift=8pt,yshift=-17.2+0.8pt] at (A.east)
	{\scriptsize 0.20};
		\node[anchor=north,xshift=8pt,yshift=-28pt] at (A.east)
	{\scriptsize 0.00};
	
			\node[anchor=north,xshift=6pt,yshift=110.5pt] at (A.east)
	{\scriptsize [mm]};
	\node[anchor=north,xshift=8pt,yshift=102pt] at (A.east)
	{\scriptsize 1.00};
	\node[anchor=north,xshift=8pt,yshift=18.4+72pt] at (A.east)
	{\scriptsize 0.00};
	\node[anchor=north,xshift=8pt,yshift=6.8+72pt] at (A.east)
	{\scriptsize 0.60};
	\node[anchor=north,xshift=8pt,yshift=-4.8+72pt] at (A.east)
	{\scriptsize 0.40};
	\node[anchor=north,xshift=8pt,yshift=-16.4+72pt] at (A.east)
	{\scriptsize 0.20};
	\node[anchor=north,xshift=8pt,yshift=-28+72pt] at (A.east)
	{\scriptsize 0.00};
	
			\node[anchor=north,xshift=6pt,yshift=-34pt] at (A.east)
{\scriptsize [mm]};
\node[anchor=north,xshift=8pt,yshift=-42pt] at (A.east)
{\scriptsize 1.00};
\node[anchor=north,xshift=8pt,yshift=18.4-72pt] at (A.east)
{\scriptsize 0.00};
\node[anchor=north,xshift=8pt,yshift=6.8-72pt] at (A.east)
{\scriptsize 0.60};
\node[anchor=north,xshift=8pt,yshift=-4.8-72pt] at (A.east)
{\scriptsize 0.40};
\node[anchor=north,xshift=8pt,yshift=-16.4-72pt] at (A.east)
{\scriptsize 0.20};
\node[anchor=north,xshift=8pt,yshift=-28-72pt] at (A.east)
{\scriptsize 0.00};

	\end{tikzpicture}

	\vspace{-3mm}
	\caption{\label{fig:ev} Error chromatogram of detected edges. The first column shows the sampled point clouds from the three models. The remaining columns show the error chromatogram corresponding to the sampled point clouds for three methods, namely, ED, MS, and HBS. The local details are enlarged on the right of each chromatogram map. These error maps share the same enlarged areas that are indicated with blue arrows.}
	\vspace{-6mm}
\end{figure*}

	\begin{table}[htbp]
	\centering
	\addtolength{\tabcolsep}{-4pt}
	\caption{\label{tab:aL1} Quantitative Comparison on edges}
	\begin{tabular}{c |c c c |c}
		\hline
		\hline
		Metrics & Accuracy(\%)  & Precision(\%) & Recall(\%) & Error($mm$) \\
			\hline
		\hline
		\multicolumn{5}{c}{\textbf{\underline{Model 1}}} \\
			\hline

		ED & 98.79  & 71.36 & 67.93& 0.3088\\%
		MS  & 99.20  & 86.08 & 73.53& 0.1233\\	
		\textbf{HBS} & \textbf{99.92}  & \textbf{98.84} & \textbf{99.17}& \textbf{0.0072}\\	
			\hline
		\hline
		\multicolumn{5}{c}{\textbf{\underline{Model 2}}} \\
		\midrule
		ED & 98.38  & 57.88 & 39.04& 0.5972\\%
		MS  & 98.94  & 74.07 & 74.56& 0.2784\\	
		\textbf{HBS} & \textbf{99.87}  & \textbf{98.07} & \textbf{97.17}& \textbf{0.0105}\\	
			\hline
		\hline
		\multicolumn{5}{c}{\textbf{\underline{Model 3}}} \\
		\midrule
		ED & 98.47  & 61.69 & 34.80& 0.5203\\%
		MS  & 98.96  & 74.08 & 73.50& 0.2673\\
		\textbf{HBS} & \textbf{99.85}  & \textbf{98.61} & \textbf{98.19}& \textbf{0.0098}\\	
		
			\hline
		\hline
	\end{tabular}
\end{table}
	
	As shown in Table~\ref{tab:aL1}, it is obvious that our baseline \we outperforms the other baselines. Notably, the complex shapes are challenging to edge detection methods, which verifies the robustness and effectiveness of our proposed method \we. The error that is introduced by the complex shapes of Models 2 and 3 increases with \{45.83\%, 36.11\%\}, while the increasing ratios are \{93.39\%, 68.49\%\} for ED and \{125.79\%, 116.79\%\} for MS. Furthermore, the average error of our method\we is approximately 0.01 mm, which meets the edge mailing of the accuracy condition in industrial manufacturing for aircraft skins.
	
	An overall visual illustration of the edge comparison of the three models is presented in Fig.~\ref{fig:ev}. The first column denotes the three point clouds that are sampled from Models 1-3. Based on this information, the last three columns from left to right denote the error chromatogram of the edge detection methods. For better visualization, we only present the color variation range from 0.00 mm to 1.00 mm and show the error that is out of the range as 'black'. The 'black' points can be regarded as points that cannot be correctly recognized by the tested edge detection method. 
	As shown in Fig.~\ref{fig:ev}(b), the edges with large eigenvalue variation are correctly detected. Most points of these edges are green, which denotes a low deviation. However, the open edges that do not have eigenvalue variation are missed by the ED method. Another line of the MS detector is shown in Fig.~\ref{fig:ev}(c). The open edges are detected with low errors. However, the closed edges show a high error that even influences the performance of neared open edges. These two typical edge detection methods cannot meet the accuracy requirements of aircraft skin milling and assembly. The error chromatograms of edges that are extracted by HBS are shown in Fig.~\ref{fig:ev}(d). In conclusion, these tests show that our proposed method outperforms and achieves fine performance in terms of accuracy, precision, recall, and error evaluation.

    \revise{\textbf{Robustness to Noise:} This experiment is designed to test the noise resistance ability of the proposed framework. We add Gaussian noise with standard deviation of 2-20 $\times10^{-3}mm$ to each point of the measured workpieces and report our results in Table~\ref{tab:Noise}. 
    As the standard deviation increases, the radius of the local search field needs to be enlarged to compensate for the uncertainty due to noise and the unavoidable loss of local details in the extracted edges. We note that even with the addition of noise, our HBS method still outperforms ED and MS.}

    \revise{In Table~\ref{tab:Noise}, we report our results for outlier experiments. We add 2-20\% number of outliers to the measured point cloud. The overall results do not change even when 20\% outliers are added. This is because our branch pruning effectively removes the outliers with the help of our graph optimization refinement and gives an accurate milling edge for practical workpiece machining.
    }

    \revise{We also consider non-uniform sampling of the parts
    and examine the influence of sparse or dense scanning. We randomly choose ten seed points and take their neighborhood area through a down-sampling or up-sampling process to change the local density of the point cloud. The results are reported in Table~\ref{tab:Noise} along with the different noise levels. We can see the performance of our method is consistent. The $\dot{x}$ represent the experiments with irregularly sampling, where $x\in \{2,4,...,20\}$. We also give the systematic assumption (3) in Section II to pre-process the point cloud by spatial filtering and ensure the efficiency of parallel computing. The experiments show that the HBS method can deal with the non-uniform sampled points.
    }

	\begin{table*}[htbp]
		\centering
		\caption{\label{tab:Noise} \revise{Noise Resistance with irregularly sampled} }
		\addtolength{\tabcolsep}{-0pt}
        \renewcommand\arraystretch{1.2} 
		\begin{tabular}{c | c c c c c c c c c c}
			\hline
			\hline
             Noise ($\times 10^{-3}mm$)        &          2         & $\dot{4}$ &6&$\dot{8}$&10 &$\dot{12}$ & 14 &$\dot{16}$ &18 &$\dot{20}$	\\
            \hline
			Error ($mm$)    	            &          0.0093         & 0.0094 &0.0096&0.0101&0.0104 &0.0108 & 0.0117 &0.0121 &0.0126 & 0.0130	\\
			\midrule
            Outlier ($\times \%$\#pts) &                2         & $\dot{4}$ &6&$\dot{8}$&10 &$\dot{12}$ & 14 &$\dot{16}$ &18 &$\dot{20}$ \\
            \hline
            Error ($mm$)    	            &          0.0092         & 0.0093 &0.0092 &0.0092 &0.0092  &0.0094 & 0.0095 &0.0094 & 0.0095 &0.0093	\\
			\hline
		\end{tabular}
		\vspace{-3mm}
	\end{table*}

	\textbf{\emph{2) Ablation Study:}} In this test, we explore the contributions with different module combinations of our proposed method, which consists of the probability detector (PD), cross validation (CV), and graph refinements (GR). These tests are designed to verify the effectiveness of these modules. The testing data are still sampled from the synthesized models, and the nonedge points (90k points) and edge points (3k points) are identical to the configurations above. Furthermore, we also add 1k points of uniform noise around the cubes that cover the whole model to increase the randomness of the ablation study. All tests were repeated ten times, and then the averages were taken as the final results. The accuracy, precision, recall, and error metrics are included for performance evaluation and are shown in Table~\ref{tab:Ablation2}. The symbol $'\checkmark'$ denotes the module selected in these combinations, and $'-'$ represents not. First, the detection only runs with the probability detector, which shows great performance but still needs to be improved. Hence, we add CV and GR to constitute the combinations on the backbone of the detector. From Line 2 of Table.~\ref{tab:Ablation2}, the PD+CV combination includes more edge points and decreases the false negatives to improve the recall ratios and contribute fewer improvements in decreasing error. This strategy indicates that our proposed method makes progress in edge point recognition to guarantee the continuity of edge lines. Different from the CV module, GR are designed to prune undesired false positives. The performance of PD+GR is shown on Line 3 of Table.~\ref{tab:Ablation2}. This module combination greatly contributes to the precision and error because the GR is the purely trimming process and does not involve new uncertainty.
	
	As shown in the last line of Table.~\ref{tab:Ablation2}, the HBS (PD+CV+GR) algorithm shows a great performance for edge detection of complex surfaces, which is often applied to aircraft skins. This combination is regarded as the standard paradigm for the robotic edge detection of our aircraft skin workpieces.

		\begin{table*}[htbp]
		\centering
		\caption{\label{tab:Ablation2} Ablation study for different module combinations} 
		\addtolength{\tabcolsep}{5pt}
		\begin{tabular}{c c c | c c c |c}
				\hline
			\hline
			PD		& CV		& GR 	 & Accuracy(\%)  & Precision(\%) & Recall(\%) & Error($mm$) \\
			\midrule
			$\checkmark	$		&			$-	$			&		$-	$				&	99.83  &  96.22 &   96.65  &  0.0214	\\
			$\checkmark	$		&			$\checkmark$	&		$-	$				&	99.87	&	96.57	 &	98.10		 &0.0187	\\
			$\checkmark	$		&			$-	$			&		$\checkmark$		&	99.84	& 	98.31	& 	97.13		& 0.0113	\\
			$\checkmark	$		&			$\checkmark$	&		$\checkmark$		&	\textbf{99.88}    &   \textbf{98.51} &   \textbf{98.18}& \textbf{0.0092}	\\

				\hline
			\hline
		\end{tabular}
		\vspace{-3mm}
	\end{table*}

    \revise{\textbf{Impact of hyper-parameters:} We perform ablation study to analyze the effect of hyper-parameters. We use precision and recall as the evaluation metrics. We also report the computational cost to highlight its tradeoff with detection accuracy.
    First, we study the effect of edge probability threshold $E_{min},E_{max}$. Table \ref{tab:Ablation6} shows our results.  Higher $E_{min}$  discards too many edge points, making the establishment of the backbone hard. 
    The subsequent extending and pruning processes are unable to recover the complete desired edges due to sparse backbones. Moreover, the large range of possibilities increases the computation time and does not improve the performance.
    Next, we study the effect of local graph radius $R_g,R_l$ values. We combine different integer multiples of the average distance $d_{\#}$ to show the impact of $R_g,R_l$ tuple. $R_g$ denotes the larger local field of graph optimization and $R_l$ is the value associated with local perturbation pruning. We report our results in Table \ref{tab:Ablation6}. Setting the values of $R_g,R_l$ to 5,3 and 10,3 $d_{\#}$, the value of $R_g$ covers the edge refinement, and a larger value increases the confidence of edge bridging between the highly confident edge patches. Setting the values to 10,3 and 10,5 show that $R_l$ determines the precision. This experiment shows the effectiveness of the local perceptive field module. Increasing $R_l$ makes the graph connect more neighborhood points and inspect more local details to prune perturbations.
    Overall, our ablation study shows that our method is not sensitive to the selection of hyper-parameters for edge detection. 
    }

		\begin{table}[htbp]
		\centering
		\caption{\label{tab:Ablation6} \revise{Impact of hyper-parameters} }
		\addtolength{\tabcolsep}{-4pt}
		\begin{tabular}{@{}cc|ccc@{}}
				\hline
			\hline
			Free parameters          	            &          Value         & Precision(\%)      & Recall(\%) & Time(s)	\\
			\midrule
            \multirow{5}{*}{$E_{min},E_{max}$}	   &			0.8,0.9			&		92.87		&  93.56   &    0.46   \\
                                                    &			0.6,0.9		&		96.16		& 
            94.78   &    0.59   \\  \cmidrule{2-5}
                                                  &			0.6,0.8			&		98.49		&    98.17      &  0.72\\
                                                    &			0.4,0.8			&		\textbf{98.51}		&   \textbf{98.18}       & \textbf{0.82}\\     &			0.2,0.8		&		98.51		&    98.18       &    3.58   \\  \cmidrule{2-5}
                                                    &			0.4,0.6				&	98.51			&    98.18     &  1.24\\ 
                                                    &			0.2,0.6			&		98.51		&         98.18  & 3.27\\
            \hline
            \multirow{5}{*}{$R_g,R_l$ (*$d_{\#}$)}	&		    $5,3$			&		94.71		&       91.10   & 0.37\\
                                                    &			$10,3$			&		96.94		    &       98.11   & 0.69\\
                                                    &			$10,5$			&		98.30		&       98.07      &  0.73\\
                                                    &			$20,5$			&		\textbf{98.51}		&       \textbf{98.18}      &  \textbf{0.82}\\
                                                    &			$40,10$			&		98.51		&       98.18      &  3.73\\

				\hline
			\hline
		\end{tabular}
		\vspace{-3mm}
	\end{table}

    \textbf{Time complexity:} We also perform an empirical study to evaluate the time complexity on different point numbers of both three models. The results are reported in Table.~\ref{tab:Ablation3} to examine the influence of the point density on our HBS method. As illustrated for the time complexity in Section~\ref{iiic}, the point number of sampling for the measured workpiece results in the different ratios of the edge point.
    The number of point cloud is sampled with $1\times 10^4,5\times 10^4,10\times 10^4,50\times 10^4,$\text{and~}$100\times 10^4$ respectively.

    As shown in Table.~\ref{tab:Ablation3}, the overall running time increasing from $1\times 10^4pts/0.13s$ to $100\times 10^4pts/0.82s$ has shown the effectiveness of the parallel hardware and knn construction accelerating. The errors are also reported as the point number variant. The low point density results in a high error. The HBS needs to enlarge the search radius and cause vague edge classification because the sparse point distribution can not support accurate STC computing. On the contrary, the edge errors achieve fine performance as the point number increases. Note that, simply increasing the point density/number is not always helpful to the desired edge detection, as shown in the last two rows of Table.~\ref{tab:Ablation3}. This experiment inspires us that the suitable sample density to reduce the time complexity but not performance in practice.

		\begin{table}[htbp]
		\centering
		\caption{\label{tab:Ablation3}  Time cost of our HBS method to different point numbers }
		\addtolength{\tabcolsep}{0pt}
		\begin{tabular}{c | c c }
				\hline
			\hline
			Point Number(\#pts)		& Running Time(s)		& Error(mm)	\\
			\midrule
            $1\times 10^4$	&			0.13			&		0.0209		\\
			$5\times 10^4$	&			0.22			&		0.0099		\\
			$10\times 10^4$		&			0.28	&		0.0093		\\
			$50\times 10^4$		&			0.51			&	\textbf{0.0092}	\\
			$100\times 10^4$	&			0.82	&		\textbf{0.0092}	\\

				\hline
			\hline
		\end{tabular}
		\vspace{-3mm}
	\end{table}
	
	\subsection{Edge measurement of aircraft skin workpieces}
\begin{figure}[t!] 
		\begin{tikzpicture}[inner sep=0pt,outer sep=0pt]
		\node[anchor=south west] (A) at (0in,0in)
		{\includegraphics[width=.48\textwidth,clip=false]{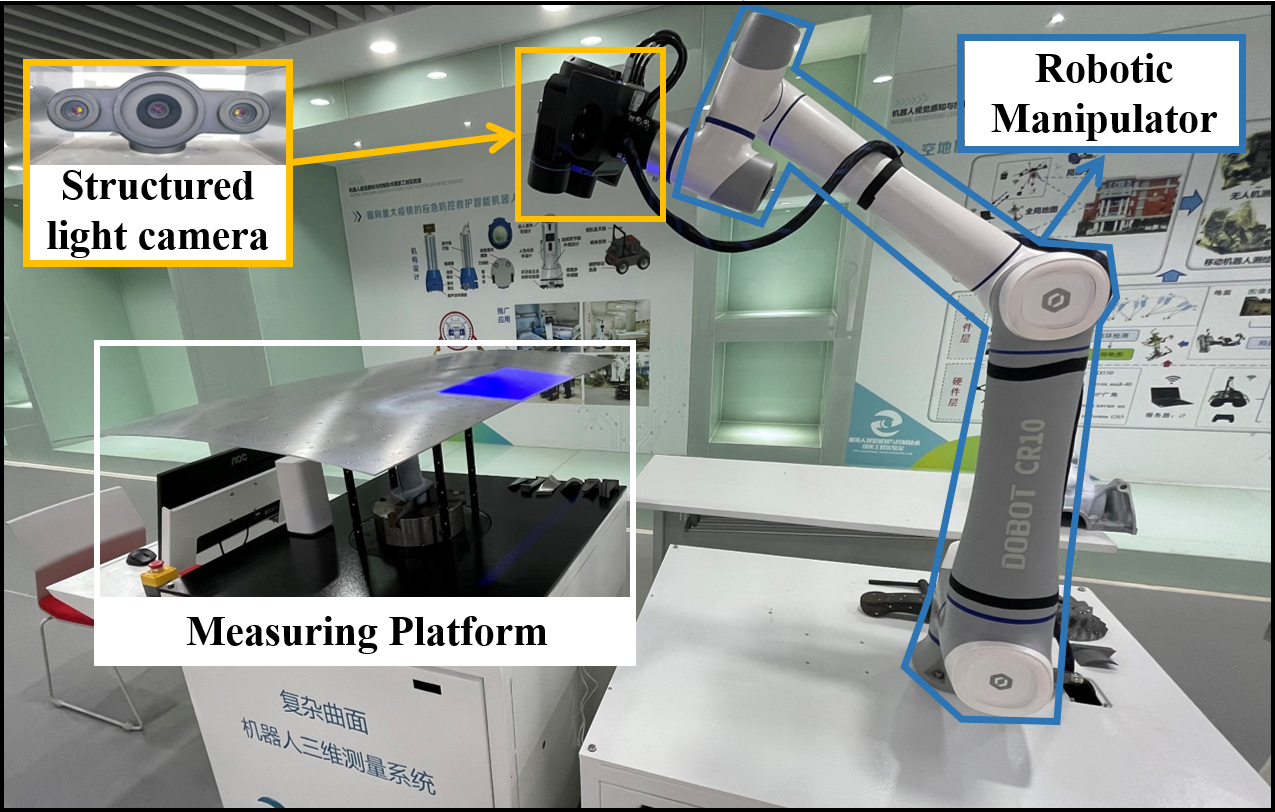}};

		\end{tikzpicture}

		\vspace{-3mm}
		\caption{\label{fig:system} Robotic measurement system.}
		\vspace{-6mm}
	\end{figure}

	In this section, we perform the edge measurement with our aircraft skin measurement system, which is introduced in Fig.~\ref{fig:system}. The whole system mainly consists of three parts: 1) structured light camera, 2) robotic manipulator, 3) the measuring platform.
    The structured light camera that is equipped at the end of a robotic manipulator measures the aircraft skins on the measuring platform. The manipulator provides different measurement positions for the view scanning of the camera. All measured aircraft skins are manufactured with the thickness of $
    \delta=5mm$. In this experiment, the controlling parameters of the manipulator are given and not considered in this article.
	
	In this experiment, we perform the edge measurements of four aircraft skin workpieces that consist of two parts: the measurement of the standard workpieces via the detected edges and edges of nonstandard workpieces visualization.
	
	\begin{figure}[t!] 
	\begin{tikzpicture}[inner sep=0pt,outer sep=0pt]
	\node[anchor=south west] (A) at (0in,0in)
	{\includegraphics[width=.48\textwidth,clip=false]{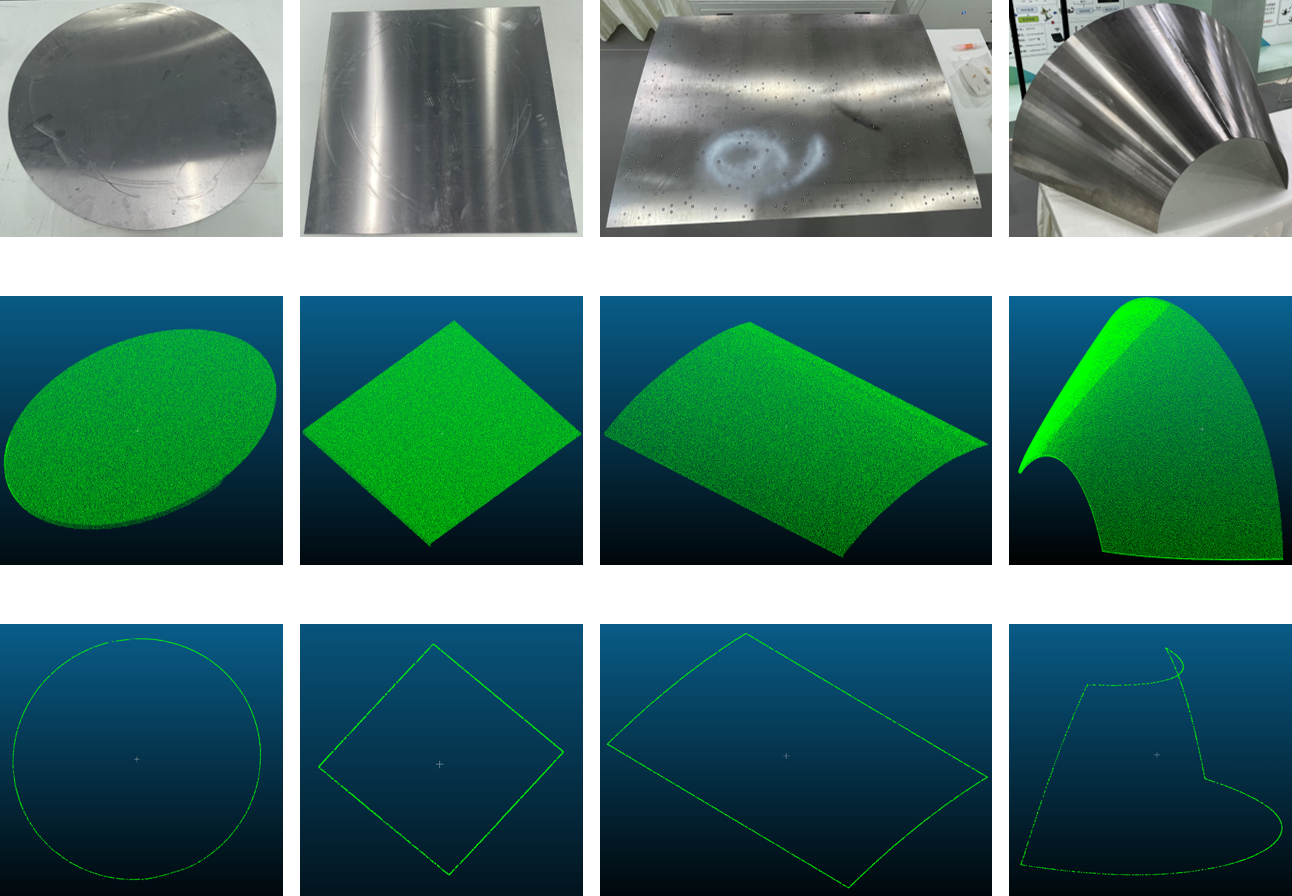}};
	\node[anchor=north,xshift=-8pt,yshift=124-60pt,rotate=90] at (A.west)
	{\footnotesize Workpieces};
	\node[anchor=north,xshift=-8pt,yshift=61-60pt,rotate=90] at (A.west)
	{\footnotesize Point clouds};
	\node[anchor=north,xshift=-8pt,yshift=-2-60pt,rotate=90] at (A.west)
	{\footnotesize Detected edges};
	
		\node[anchor=north,xshift=0pt,yshift=124pt] at (A.south)
	{\footnotesize (a)};
	\node[anchor=north,xshift=0pt,yshift=61pt] at (A.south)
	{\footnotesize (b)};
	\node[anchor=north,xshift=0pt,yshift=-2pt] at (A.south)
	{\footnotesize (c)};
	
	\node[anchor=north,xshift=-98pt,yshift=180pt] at (A.south)
	{\footnotesize Skin 1};
		\node[anchor=north,xshift=-98+60pt,yshift=180pt] at (A.south)
	{\footnotesize Skin 2};
		\node[anchor=north,xshift=-98+60+65pt,yshift=180pt] at (A.south)
	{\footnotesize Skin 3};
		\node[anchor=north,xshift=-98+120+75pt,yshift=180pt] at (A.south)
	{\footnotesize Skin 4};
	
	\end{tikzpicture}

	\vspace{-3mm}
	\caption{\label{fig:me} Edge Detection using HBS on Skin Models.}
	\vspace{-7mm}
\end{figure}

	\textbf{\emph{1) Standard workpieces measurement:}} Here, we apply our proposed HBS method to measure the parameters via two standard workpieces. These two workpieces (skin 1, 2) are designed as a standard circle with a radius of 0.25$m$ (skin a) and a square with a size of 0.5$m$$\times$0.5$m$ (skin b). The thickness is designed as 5 $mm$ to approach the size of aircraft skins. The point clouds of the skins are scanned with our robotic measurement system, and the path of the scanning manipulator is pregiven.
	
	We apply the coordinate average distance error, distance $d_{z}$ along the z-axis that indicating flatness, and standard deviation $Std$ as the evaluation metric. The point sets are manually aligned to the standard parameters. These performance evaluation details are shown in Table~\ref{tab:cs}. Here, the comparisons are presented in this test. Because the EV cannot simultaneously detect the open edge and closed edge, we increase the scanning times (abbreviated as S.t.) to include the surface thickness and ensure that all desired milling edges are scanned as closed edges. As shown in the circle comparison of Table~\ref{tab:cs}, the EV method demands nearly twice the other methods to approach the desired edges but has greater performance than the MS method. It can be concluded that the curvature and eigenvalue of all upper edges are similar in circle skins. Unlike the performance in Section.~\ref{iiia}, the performance of the MS method is limited by the multiple edge types that were introduced during scanning. Our HBS outperforms the baselines and achieves fine milling edge detection. In the square comparison of Table~\ref{tab:cs}, there is edge variation, such as the corners and lines. EV shows a significant error increase but still outperforms the MS method due to the greater number of scans. In contrast, MS and HBS still require few scans. HBS is indicated to be suitable for edge detection for automatic machining. Different from the current edge detection methods, HBS is designed to detect the upper milling edges and is experimentally shown to be superior.

	\begin{table}[htbp]
		\centering
		\caption{\label{tab:cs}Performance comparison on skins}
		\addtolength{\tabcolsep}{0pt}
		\begin{tabular}{c c c c c }
			\hline
			\hline
			Circle		& $Error$($mm$)		&$d_{z}$($mm$)	 &$Std$($mm$)&S.t.\\
			\hline
			EV		&			0.1124			&		0.0678		  &	  0.1059   &  23 times \\
			MS			&		0.4912	&		2.7123	      &	0.4219	& 11 times	\\
			HBS	&		\textbf{	0.0112}	&		\textbf{	0.0096	}       &	\textbf{	0.0173}	& 11 times	\\
				\hline
			\hline
			Square		& $Error$($mm$)		&$d_{z}$($mm$) &$Std$($mm$)	&S.t.\\
				\hline

			EV		&			0.2015			&		0.0893	  &	  0.2876     &   25 times\\
			MS			&			0.5789	&		3.2310      &	0.5134	& 13 times	\\
			HBS	&		\textbf{	0.0104}	&	\textbf{	0.0086 }  &\textbf{	0.0141}	& 13 times	\\
			\hline
		\hline
		\end{tabular}
	\end{table}
	
	
	
	\textbf{\emph{2) Visualization of Measured Workpieces:}} Here, we apply our \we method to detect the edge on two nonstandard aircraft skins and then visualize the detected edges, including open and closed edges, which are detailed in Fig.~\ref{fig:me} to show the effectiveness of our method. The overall illustration shows that HBS is suitable for performing edge detection for further milling allowance tasks.

\begin{figure}[t!] 
		\begin{tikzpicture}[inner sep=0pt,outer sep=0pt]
		\node[anchor=south west] (A) at (0in,0in)
		{\includegraphics[width=.48\textwidth,clip=false]{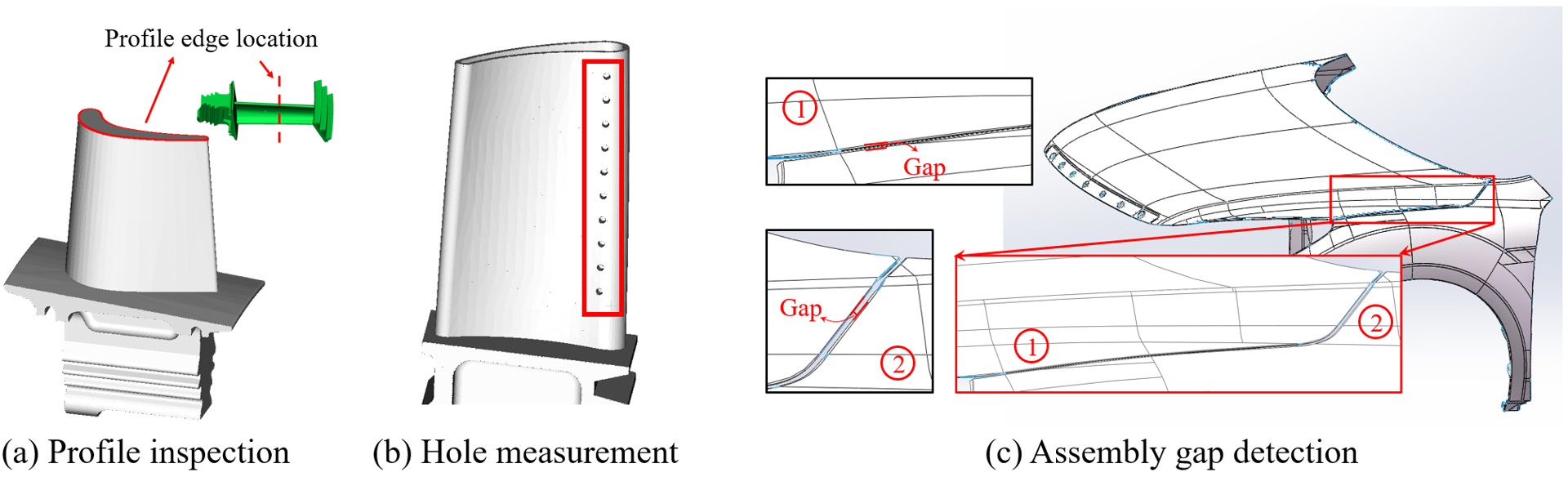}};

		\end{tikzpicture}

		\vspace{-3mm}
		\caption{\label{fig:more} \revise{Edge detection in diversified scenarios using HBS.}}
		\vspace{-6mm}
	\end{figure}

    \revise{\textbf{\emph{3) Extended Applications:}} We show the generalization ability of our HBS method and test it in various extended applications as shown in Fig.~\ref{fig:more}, including blade profile inspection, hole measurement, and assembly gap detection. Given the CAD models for measurement, we simulate the view scan process using Blender Sensor Simulation (BlenSor~\cite{blensor}). We report the measurement results using HBS in Table~\ref{tab:profile_all}.
    \begin{enumerate}
        \item \textbf{Profile inspection:} We sample the profile of the blades that can be used as the quality inspection for aviation manufacturing. The edge of the profile is extracted by HBS and then evaluated with the metric, same as the skin performance comparison. Furthermore, we evaluate the flatness $d_{z}$ of profile surface using the detected edge.
        The details of the profile edge are reported in Table~\ref{tab:profile_all}. The overall performance meets the condition of profile measurement.
        \item \textbf{Hole diameter measurement} can be used in many diversified scenarios, such as drilling and riveting, and surface hole measurement. We first apply our view sample simulation to get the surface point cloud around the target holes. Note that for multiple circle detection, the linear degree $\theta$ is increased to allow distortions of circle edge and hence different to our skin milling edge detection. $\theta=1$\degree here. The detailed diameter computing is the same as our circle milling edge detection in Table.~\ref{tab:cs}. The measured average diameter is 1.5014 which achieves an accurate hole diameter measurement compared to the ground-truth (GT) 1.5000.
        \item \textbf{Assembly gap detection} is also an important application using edge detection. Assembly quality can be computed by the detected edges of the adjacent workpieces. We get point clouds from the assembly position to get the two edges from the adjacent car parts respectively. The distance to the nearest neighbor point (1-NN) searched from another edge is regarded as the coarse gap and the average gap is reported in Table~\ref{tab:profile_all}. We use the same metrics as the former edge point evaluation. This example of car assembly gap measurement shows that the edge detection technique is an essential process for automatic measurement.
    \end{enumerate}} 

    \begin{table}[htbp]
		\centering
		\caption{\label{tab:profile_all} \revise{Three Extended Applications using HBS}}
		\addtolength{\tabcolsep}{-6pt}
		\begin{tabular}{c | c c c }
			\hline
			\hline
			Metric		& $Error$($mm$)	& $Std$($mm$)& Flatness $d_{z}$($mm$)\\
			\hline
			\tabincell{c}{Profile \\inspection}	&		0.0013	      & 0.0016 & 0.0009	 \\ 
            \hline
   		Metric		& $Error$($mm$)		&$Std$($mm$)	 & GT/Avg.Diameter($mm$) \\
			\hline
			\tabincell{c}{ Hole diameter \\ measurement}	& 0.0016	&		0.0007 & 1.5000/\textbf{1.5014}	\\
				\hline
    		Metric		& $Error$($mm$)		& $Std$($mm$)	& 1-NN Gap($mm$)\\
			\hline
			\tabincell{c}{Assembly \\gap detection}	&		0.0011	&		0.0014 & \textbf{4.0010}    	\\
		\hline
        \hline
		\end{tabular}
	\end{table}

    \section{Discussion}
    The current edge milling scheme is only the tip of the iceberg for industrial measurement. It is crucial to fuse different requirements like skin machining and planning to integrate our proposed techniques into the whole automatic industrial manufacturing.
    Our method has several limitations. First, it is not designed for the separate detection of feature lines, so the detailed measurement would be miss out even if the edges have been well-detected. In the future, we plan to investigate how to enhance the measurement framework and the edge graph for more detailed measurements of aircraft skins, such as size inspection and quality evaluation. Besides, with insufficient points, the measurement framework needs to enlarge the neighbor searching radius and hence ignore some key points. Moreover, the current implementation considers point patches with a fixed radius of the perceptive field instead of adaptive structures with varying scales. We plan to design an adaptive structures and employ the novel theories to achieve more accurate aircraft skin measurement in our future works.

	\section{Conclusion}
	In this paper, we proposed a novel edge detection method for the edge machining of aircraft skins, which was introduced for accurate upper edge extraction with the scanning of complex skin structures. Our proposed method, which consists of the initial backbone establishment, edge extension, and refinement modules, accurately realizes milling edge detection. The detected upper edge is employed in automatic aircraft skin assembly and aviation manufacturing. According to the comparisons between our proposed method and other baselines, our approach performed well and presented a novel paradigm for the milling edge detection task.

	\appendices
	%


	\ifCLASSOPTIONcaptionsoff
	\newpage
	\fi
	
	\bibliographystyle{IEEEtran}
	\bibliography{Reference}
\end{document}